**Computer Vision and Its Relationship to Cognitive Science: A perspective from Bayes Decision Theory.** *Alan Yuille and Dan Kersten*.

This document presents an introduction to computer vision, and its relationship to Cognitive Science, from the perspective of Bayes Decision Theory (Berger 1985). Computer vision is a vast and complex field, so this overview has a narrow scope and provides a theoretical lens which captures many key concepts. BDT is rich enough to include two different approaches: (i) the Bayesian viewpoint, which gives a conceptually attractive framework for vision with concepts that resonate with Cognitive Science (Griffiths et al., 2024), and (ii) the Deep Neural Network approach whose successes in the real world have made Computer Vision into a trillion-dollar industry and which is motivated by the hierarchical structure of the visual ventral stream. The BDT framework relates and captures the strengths and weakness of these two approaches and, by discussing the limitations of BDT, points the way to how they can be combined in a richer framework.

**Vision Systems and Human Perception**

The goal of a vision system is to infer information about the world from images captured by cameras or human eyes. This is a profoundly difficult task. The human ability to interpret visual data has been called our underappreciated superpower. Cognitive science seeks to understand how humans accomplish this, while computer vision attempts to develop vision systems that perform this task in complex real-world environments. Both fields began to flourish in the late 1970s, enabled by growing access to computational resources for exploring human and artificial intelligence. Both built on the long history of psychophysics of human perception and the growing understanding of the biology of vision, and particularly the hierarchical structure of the visual cortex. Computer vision researchers developed theories which perform well on complex real images which required borrowing, adapting, and exploiting a huge variety of technical tools from a range of disciplines including signal processing, linear algebra, algorithms, pattern recognition, cybernetics, statistics, physics, and geometry.

**Cognitive Science and Computer Vision: Three Levels of Explanation**

Marr and Poggio (Marr 1982) speculated about the relationship between biological vision systems and computer vision, and proposed a framework consisting of three levels of analysis:

1. Computational level – where both human and machine vision share similar goals of obtaining information about the world from images.

2. Algorithmic level – the specific algorithms used by AI and humans, which may differ between biological and artificial systems.

3. Implementation level – where biological "wetware" and digital hardware are fundamentally distinct.

This framework enabled researchers to study vision at different levels of analysis. The computational level was particularly attractive to those vision researchers who wanted to develop theories for both human and computer vision, but who did not want to deal with the underlying biology. Most computer vision researchers, however, had limited interest in human perception. This distinction between computational and algorithmic levels became blurred as computer vision researchers increasing relied on neural networks.

This level of explanation framework does not imply that computer vision researchers should develop systems that mimic all aspects of human vision. While human vision remains the gold standard for many visual tasks, there are situations where computer vision systems now outperform it. Biological systems are constrained by limited memory and computational power and were developed by evolution to perform visual tasks of value to human survival and not, for example, to detect tumors in computed tomography images. There are also many visual phenomena including inattentional effects like failing to see a gorilla in the room due to attending elsewhere, where humans make mistakes which computer vision would not. It is conjectured that the human visual system uses strategies which are energy efficient and have evolved to perform those visual tasks relevant to our ancestors.

**The Bayesian Perspective on Vision**

Early vision scientists like Helmholtz and Gregory anticipated Bayes by showing that the human visual system combines both direct sensory processing with prior knowledge about the world stored in memory.

More precisely, Bayes formulates vision as an inverse inference problem whose goal is to infer the state of the world $W$ from an image $I$. Two key concepts are the likelihood function $P(I|W)$ for generating the image and the prior probability of the world state $P(W)$. This can be thought of as inverse-computer-graphics if the generative process $P(I|W)$ is specified by a graphics engine. The likelihood function alone is not sufficient to determine the world's state uniquely and prior knowledge is required to obtain a unique percept. Bayes advocates selecting the $W$ that maximizes the posterior probability $P(W|I) = P(I|W) P(W)/P(I)$, which combines evidence from the likelihood and the prior. This is illustrated in Figure 1 where the likelihood alone is not sufficient to give an unambiguous percept, and a prior in shape is required supplemented by assumption that the object is not seen from an unusual viewpoint (Knill and Richards Chp 9).

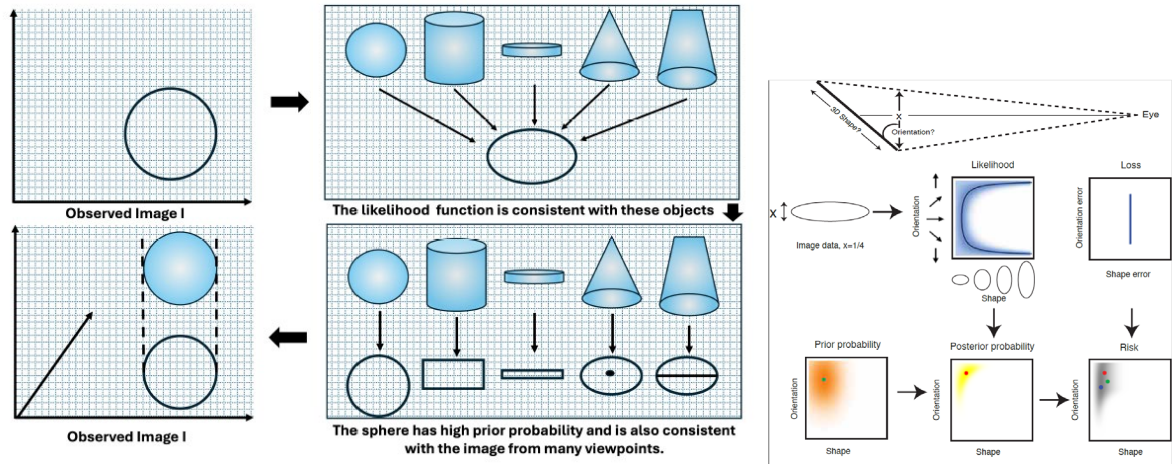

Figure 1: The likelihood function P(I|W) puts restrictions on which objects can generate an image but is not sufficient to yield a unique precept. This requires making additional prior assumptions P(W) about the objects. Left Panel: the image of a circle is consistent with the projection of several objects but of these, the sphere has high prior probability and has similar appearance from many viewpoints. Right panel: if instead we interpret the image as a planar two-dimensional object, then the likelihood, black curve, is insufficient and a prior is required for a unique percept. The blue gradient represents degrees of uncertainty due to noise in the data. The prior is that surfaces tend to be oriented away from the viewer and have shapes closer to a circle. The loss function prioritizes accurate estimation of the 3D shape compared to orientation. The green, red, and blue dots mark the maximum prior, maximum posterior, and minimum risk, respectively.

Many visual demonstrations suggest that humans are capable of approximate inverse-computer-graphics consistent with Bayes. For example, Michael Bach's webpage https://michaelbach.de/ot/ provides many visual illusions with Bayesian interpretations. Human illusions of "flying carpets" and "levitations" of images, see Figure 2, are hard to explain without assuming that humans are attempting to estimate the 3D world that generated the image and, in particular the positions of shadows. This can be quantified in experimental settings like Kersten's "ball in a box" shown in
https://kerstenlab.psych.umn.edu/demos/shadows/

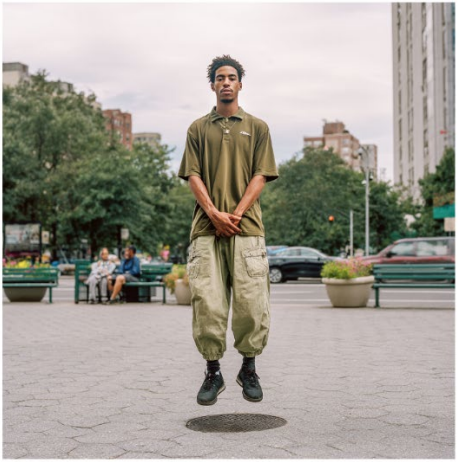
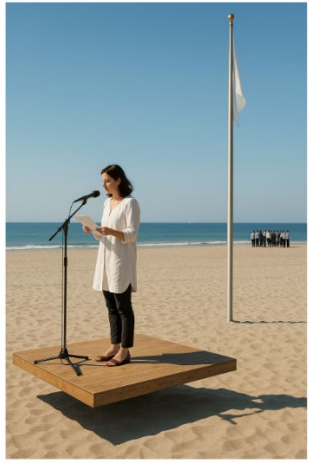

*Figure 2: Illusions like levitation and flying carpets demonstrate that human perception tries to find 3D explanations of visual stimuli using shadows and other cues. The illusions arise due to accidental alignments of shadows and other dark image regions.*

The Bayesian perspective acknowledges that inverse-computer-graphics is often unnecessary for computer vision and biological vision systems. For many tasks, like playing baseball or driving a car, a vision system only needs to extract information which is relevant for the task. But the demonstrations suggest that humans can perform inverse-computer-graphics approximately and as-needed. It has been hypothesized that this can be performed by a bottom-up and top-down process exploiting the structure of the visual cortex and the huge number of top-down connections (Mumford 1992).

**Modularity**

The complexity of vision motivated researchers to break it down into modular components that could be studied separately (Marr 1982). Classic examples, illustrated in Figure 3, include segmentation, depth estimation, and object recognition. These can be decomposed even further into depth from binocular stereo, structure from motion, shape from shading, and shape from texture.

Modularity makes it practical to develop Bayesian theories for some vision modules by designing plausible likelihoods and priors. But until recently it was not practical to do this for object recognition, let alone for inverse-computer-graphics. Even when annotated datasets became available researchers lacked the tools to learn these probabilities and instead learnt the posterior distributions by feedforward deep networks as discussed later.

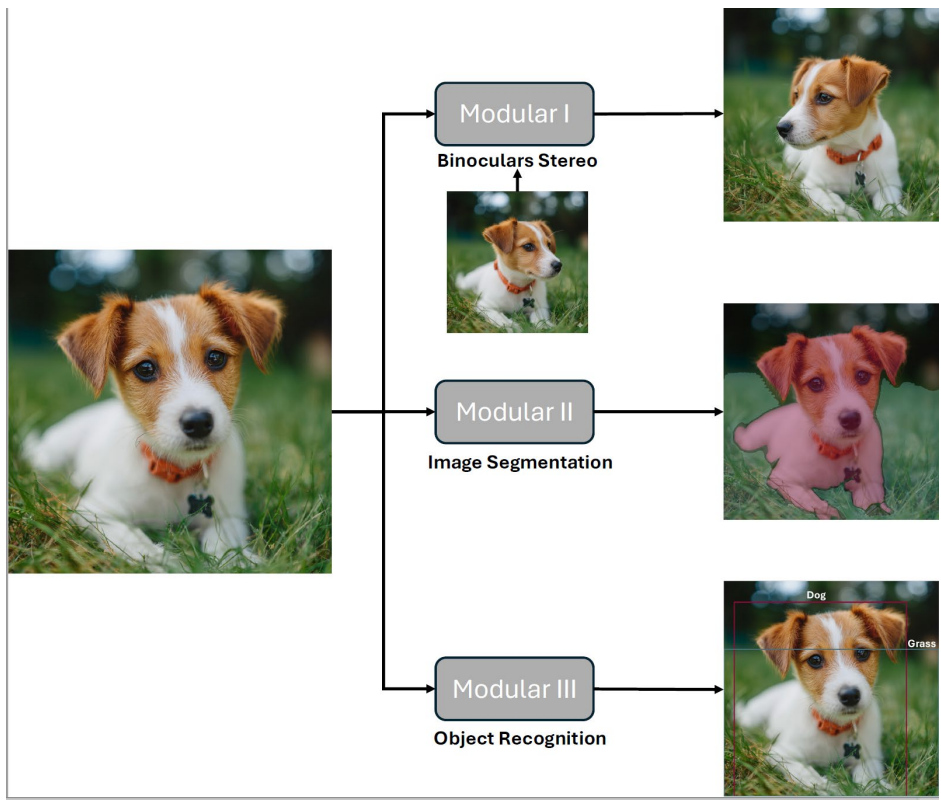

*Figure 3. Vision can be subdivided into a set of tasks performed by vision modules. Important modules include: (I) Binocular Stereo, where depth is estimated from two eyes. (II) Image Segmentation, which identifies and estimates the boundaries between a foreground and a background object. (III) Object Recognition, which specifies the name of a viewed object.*

**Modules: Binocular Stereo, Motion, and Structure from Motion**

Early computer vision research focused on modular tasks like depth from binocular stereo, optical flow, and recovering 3D structure from motion. These are special cases of obtaining depth from multiple image frames.

Analyzing these tasks revealed a major challenge: the correspondence problem, where pixels must be matched across image frames. This correspondence problem is ambiguous and requires assumptions about the structure of the world, such as that surfaces are piecewise smooth, and objects typically move slowly and smoothly.

Early researchers (Marr & Poggio 1976) formulated the stereo correspondence problem by an algorithm which, with hindsight, minimizes an energy function containing two terms. The first term depended on the input images and required that pixels in the left image were matched to pixels in the right image with similar features. The second term-imposed smoothness on the correspondence which translated into smoothness of the depth.

Similarly models for optical flow (motion in the image) assume that the flow is slow (Ullman 1979) and, more generally, slow and smooth (Yuille and Grzywacz 1987). After correspondence had been found the shape of an object could be estimated assuming that it was moving rigidly (Ullman 1979). Efficient linear algebra methods were developed to estimate object shape (Tomasi and Kanade 1993).

These energy function theories could be aligned with Bayesian formulations via Gibbs distributions as described in (Clark and Yuille 1990) where probability distributions are expressed by the exponential of the negative of the energy function. The first and second energy terms corresponded to the likelihood and the prior respectively. The priors on spatial properties could be expressed in terms of Markov Random Fields (MRFs). Minimizing the energy is equivalent to maximizing the probability, and hence to Bayesian inference. These theories could be explicitly reformulated as Bayesian, as illustrated in (Knill & Richards Chp8 1996).

**Image Segmentation**

Image segmentation is the task of decomposing images into different regions which ideally correspond to objects. An influential explicitly Bayesian model was developed for this task using the assumption that images were smooth within each region and discontinuous at the boundaries (Geman and Geman 1984). This restricted assumption could be relaxed in later models by, for example, assuming that the image regions are generated by distinct probability distributions (Zhu and Yuille, 1996).

**Shape from Shading and Texture**

Researchers also studied the problems of estimating the 3D shape of isolated objects either from shading or texture patterns. Both were formulated as inverse-computer-graphics. For shape from shading, the objects were restricted to having constant texture and Lambertian reflection were used to model how the image intensity depended on the shape of the object and the illumination. Estimating the shape of the object was formulated as energy minimization, which could be reinterpreted as Bayes using a prior that the shape was piecewise smooth (Ikeuchi & Horn 1981).

For the related task of shape from texture, the theories first filtered out the shading so that the image depends only on the projection of the texture patterns on the objects. The shape of the object could be estimated by assuming that the texture was evenly distributed on an object's surface (Blake and Marinos 1990, Knill and Richards, Chp 7. 1996.)

**Coupling Visual Cues**

Bayes gives a conceptual framework to describe how multiple visual cues -- like stereo, shading, and motion – could be combined optimally by considering the uncertainty of each cue and their statistical dependencies (Clark and Yuille 1990). An influential theory (Marr 1982) proposed that visual cues were implemented modularly, and their results combined in representations like the 2 1/2D sketch. Quantitative studies showed that some cues could be combined by weighted averaging (M. Landy et al. 1996). Bayes distinguished between weak coupling, where the cues are statistically independent, and strong coupling where they are not. It showed that weighted linear averaging arose as a special case which assumed independence and Gaussian probabilities. Bayes was able to account for psychophysics findings when shading and texture were combined (H. Buelthoff and Mallot 1988) on different types of texture cues (K. Doya et al. Chp 9. 2007) and a range of other phenomena investigated by the nascent field of computer graphics psychophysics (Knill and Richards 1996, Ernst and Banks 2002).

**Inference Algorithms for Bayesian Theories**

Bayes does not specify inference algorithms to estimate the most probable state of the posterior distribution (MAP). But for several vision modules there are algorithms which are not only effective for approximately estimating MAP but which are also neuronally plausible.  The simplest was the use of steepest decent for shape from shading (Ikeuchi and Horn 1981).  For others the context prior of weak spatial smoothness could modeled by Markov Random Fields (MRFs) (Geman and Geman 1984, Clark and Yuille 1990).  For this class of models a novel algorithm, known as mean field theory (MFT) adapted from Statistical Physics, could perform approximate inference for a range of computer vision tasks (Clark & Yuille 1990).  Examples included a neuronally plausible algorithm for the early visual cortex (Koch et al. 1986), which were later found to be roughly consistent with some neuroscience studies (K. Doya et al. Chp 8. 2027).  In addition, researchers developed faster learning based inference algorithms, but these are beyond the scope of this article.

**Bayes and its Extensions.**

Bayes can be naturally extended to combine information over time for tasks like tracking objects (Blake & Yuille. Chps.1989) and, in particular,  by Bayes-Kalman filters (Isard and Blake 1998).  It can also be extended to enable an agent to control a system by making a sequence of decisions. This was exemplified by the seminal work by Dickmanns and his collaborators on dynamic scene understanding, which resulted in automated cars that could drive 1,000 kilometers on the autobahn (Blake and Yuille. Chp 18. 1992). The vision component detected intensity edges from images, corresponding to the boundaries of the roads and cars. It applied decision and control theory to make optimal choices for how to steer the car, to keep it in lane, and to avoid other cars.

**Object Recognition and High Level Vision**

High level vision tasks including object recognition were typically studied on simplified images of edges and features. In particular, Ullman (1996) formulated object recognition as aligning stored 3D object models with features in the 2D image which is consistent with Bayes. He describes core problems, such as dealing with viewpoint, and proposed inference algorithms including a counter-stream approach related to analysis by synthesis. This work describes other high level vision problems, contains psychophysical studies, and describes earlier work such as representing 3D objects by generalized cylinders (Marr 1982) and geons (Biederman 1987). An explicit Bayesian approach to object recognition was formulated in (Liu et al. 1995) which showed that models with 3D object representations were more consistent with human performance than alternatives with 2D representations.

Bayesian models for object recognition were not developed for real images because, it is only recently becoming practical to specify realistic likelihood functions. A notable exception were the SMPL models (Loper et al. 2023), which can represent the shape and pose of humans, and can estimate the shape by predicting the boundaries of the objects and the positions of key points (see later section). These classes of models can be used to model the experimental finding of Johannsen (Johansson 1973) on biological motion.

**Bayesian Decision Theory (BDT)**

Bayes is a special case of Bayes Decision Theory (Berger 1985). BDT, as illustrated in Figure 4, is a normative approach to making decisions under uncertainty which, in addition to probability distribution P(I,W) introduces a decision rule and a loss function for making the wrong decision. It proposes that the decision rule should be chosen to minimize the risk, or the expectation of the loss function with respect to the probability of the data. If the loss function penalizes all errors equally, then the best decision rule is the MAP as in the Bayesian theories above. In practice, loss functions are needed because not all losses are equally costly. The MAP estimator is problematic when estimating a continuous variable because it corresponds to penalizing all errors equally, whether they are infinitesimal or enormous, which can be corrected by a loss function (Knill & Richards Chp. 4. 1996).

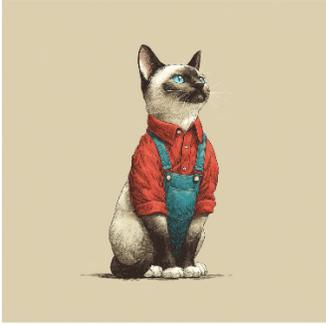

Bayesian Risk in Image Classification Example *("cat" vs "dog")*

$$R(\alpha) = E_{W,I}[L(W, \alpha(I))] = \sum_I \sum_W L(W, \alpha(I)) p(W, I)$$

$$R(\alpha \mid I) = E_{W\mid I}[L(W, \alpha(I))] = \sum_W L(W, \alpha(I)) p(W \mid I)$$

$I$ is the Observed Image,
$\alpha$ is Model Prediction ("cat" vs "dog"),
$W$ is Ground Truth Label

In Empirical Bayes, we have

$$R(\alpha \mid I) \approx \frac{1}{n}\sum_{i=1}^{n} \sum_W L(W, \alpha(I_i)) p(W \mid I_i)$$

FIGURE 4: *Bayes Decision Theory. The risk is the expected loss with respect to the distribution of W and I. The empirical risk is the average loss over data samples*.

BDT assumes that we know the probability distributions, but in practice all we have are data which can be exploited in several ways. The first is to use it to learn the probability distributions P(I|W) and P(W). Researchers learnt probability distributions for natural images and for textures (Mumford and Desolneux. Chp 4. 2010). A Bayesian theory for edge detection learnt the probability distribution for feature responses on edges and in the background (Konishi et al. 2003).

A second approach uses the data to directly learn the decision rules, instead of learning the probabilities. The learning minimized the average loss over the data samples, known as the empirical risk. An influential example was the detection of faces in real images (Viola and Jones 2001) which combined the Machine Learning techniques of AdaBoost and Decision Trees. Another example, successful for object classification, were deformable part models (Felzenszwalb et al. 2008) which used structured Support Vectors machines with a hierarchical architecture. Neural networks often were trained to learn the conditional distribution P(W|I) which is a special case.

From the BDT perspective, the optimal decision rule depends only on the posterior distribution P(W|I) so arguably there is no need to learn the likelihood and the prior. A counterargument is that if data is limited it may be sometimes easier to model the prior and likelihood, as illustrated by the vision modules described earlier, and by many cognitive examples (Griffiths et al. 2024). Moreover, as illustrated in (Konishi et al. 2003) the Bayesian approach can make it easy for an algorithm trained on one type of data (images of house interiors) to generalize and perform well on other data (images of the countryside).

**Datasets: Testing and Benchmarking**

Computer vision algorithms are evaluated, and compared to alternative theories, by their average loss on testing datasets which consists of data from the same underlying source. In general algorithms struggle to generalize and perform well on data that comes from another source which differs significantly from the training data.

Vision scientists evaluate theories in a similar way. The decision rule is replaced by the prediction of the theory and the loss function measures the difference between the prediction and the response of a human subject. One significant difference is that vision scientists use experimental design and test their theories on data from a variety of controlled sources. For example, a theory of object recognition may be tested on two different datasets where the first emphasizes texture cues and the second de-emphasizes texture and enhances shape.

**Ideal Observer Theory**

Human perception can also be tested using BDT by comparing human performance to that of an Ideal Observer (Knill and Richards 1986, Liu et al. 1995, Geisler 2011) who minimizes the Bayes risk. The data is generated from known distributions P(I,W) and P(W). In most cases, the Ideal Observer performed much better than humans partly because: (i) it knows how the data is generated, which it can exploit when making its decision, and (ii) human vision is adapted for making decisions in natural situations and not in laboratory experiments. Typical Ideal Observers lack realism by using simple models for generating images, enabling the computation of Bayes Risk, because realistic models were unavailable.

The success of deep networks means that we can learn "approximate ideal observers" by training the networks on datasets. These are only approximate because the deep networks are not guaranteed to achieve Bayes Risk (researchers may develop a better deep network). For many tasks, humans outperform these observers, suggesting that they are too approximate. See (Kell and McDermott 2019) for more on this topic and how it can be applied to models of the ventral stream.

**Features, Hierarchical Features, and Deep Networks**

Features have a long history in computer vision because they give a straightforward way to process images. For example, edge detection could be performed by convolving an image with a derivative filter and thresholding the result. A filterbank of Gabor filters could be used to classify textures or to provide local descriptions of images which could be used to address the correspondence problem in binocular stereo and motion.

Studies of the mammalian ventral stream showed it has a hierarchical structure where the low level filters were similar to Gabors and neurons higher up the hierarchy were more specific to image patterns and less tuned to their position in the image. This inspired computational models of the visual cortex (Fukushima 1980) which became increasingly realistic (Riesenhuber & Poggio 1999). Multi-layer perceptrons, whose weights could be learnt, were modified to create convolutional neural networks (CNNs) (LeCun et al. 1989), which foreshadow modern deep networks. From the BDT perspective, these neural networks specify a class of decision rules parameterized by the weights of the networks. The risk of decision rules are differentiable functions of the weights so the best decision rule can be learnt by stochastic gradient descent which is called backpropagation by how it exploits the hierarchical structure.

**Convolutional Neural Networks and Deep Learning**

CNNs became very successful due to the availability of large, annotated datasets and GPUs for efficient computation. CV researchers were skeptical of CNNs because they had been outperformed by deformable part models on smaller datasets. This changed with the success of AlexNet (Krizhevsky et al. 2012) on object classification on ImageNet , partially due to their computational efficiency including shared features. This stimulated research on neural network architectures, for instance ResNet (Kaiming He et al. 2016) which introduced residual connections, making it possible to train networks with enormous number of layers and resulting in better performance.

CNNs were rapidly applied to all visual tasks for which large, annotated datasets were available. This resulted in enormous increases in performance on tasks like: (i) semantic segmentation (Long et al. 2015) whose task is to segment images into distinct semantic regions, (ii) edge detection (Xie & Tu 2015), and (iii) 2D pose estimation whose goal was to detect the joints of humans (Chen & Yuille 2014).

These successes inspired researchers to create datasets and develop novel neural network architectures for vision tasks that had hitherto been considered intractable. Notable examples included : (i) action recognition, which introduced a network that combined image and motion cues (Simonyan & Zisserman 2014)., and (ii) generating a text caption from an image by a network which combined vision and language features (Karpathy & Fei-Fei 2015).

**Transformer Neural Networks**

The success of CNNs was challenged by Transformers (Dosovitskiy et al. 2021) which was a neural architecture developed for Natural Language tasks. Their ability to model long range context adaptively has been critical to the success of Generative Pre-trained Transformers (GPTs). This was achieved by self- and cross-attention mechanisms that enable dynamic binding between neurons allowing long-range interactions to exploit spatial context. This contrasts with CNNs where neurons have fixed sized receptive fields that capture only local and non-adaptive context. Hence transformers have some ability to perform sideways interactions characteristic of MRFs. Other attractive properties of transformers include their ability to partially segment an object despite only being trained to classify it. Intriguingly they share some of the attractive properties of grammatical approaches to modeling image patterns as compositions of basic elements (D. Mumford and A. Desolneux Chp 3. 2010) while being simpler to implement. Transformers have increasingly replaced CNNs for most vision tasks, sometimes with hybrid architectures which combine attention mechanisms with convolutional filters.

**The Rise of Foundation Models**

Foundation models were developed because researchers noticed that the performance of networks trained on data from one source degraded when tested on data from another source (e.g., object detectors trained on ImageNet make mistakes when used by automated cars). Their performance is inflated because they can exploit biases in the datasets they are trained on, analogous to how Ideal Observers can outperform humans. A related problem is that networks could be fooled by minor perturbations to images which were often invisible to humans (Szegedy et al. 2013). From the BDT perspective, both problems arise because the datasets are not big enough to represent the complexity of natural images.

Foundation Models address these problems using several related strategies. Firstly, researchers developed unsupervised learning algorithms to learn neural network features from enormous numbers of unannotated images. Particularly successful methods were iBOT and DINOv2 which combined masked-image-modeling and contrastive learning (J. Zhou et al. ICLR 2021, M. Oquab 2021). These learnt features resulted in a *backbone that could be shared for many visual tasks requiring only* small decision *heads* which were trained for each task. Secondly, researchers created enormous, annotated datasets using an active learning strategy where networks trained on annotated datasets predicted *pseudo-labels* for unannotated images which were checked and, if necessary corrected, by human annotators and then used to retrain the network. An example is Segment Anything (SAM) for segmenting objects in images by learning to exploit the cues that the texture within an object typically differs from the texture in the background and there is often intensity discontinuities at the boundary (Kirilov et al. 2023) . A third strategy is to train networks using simulated datasets, like those created for video games, for which enormous amounts are available, as exemplified by Depth Anything which estimates the shape and depths of objects from single images (L. Yang et al. NeurIPS 2024). These foundational models are very exciting because they suggest that there is critical size when datasets, for some visual tasks, become so large that the algorithms work on almost all real-world images. From our BDT perspective, this suggests that there is no need for priors and likelihoods because networks can be trained to directly model the posterior.

**Vision Language Models and Large Language Models**

The success of neural network models for text captioning encouraged researchers to develop models that combined vision and language, enabling vision researchers to benefit from the ability of Large Language Models (LLMs) to capture common sense knowledge about the world (Devlin et al. 2019). This required algorithms like CLIP (Radford et al. 2021) for converting image features to text features followed by training transformer networks on paired images and text captions. It results in Vision Language Models (VLMs) which perform amazingly well on a range of tasks including vision question and answering (VQA), by exploiting commonsense world knowledge, but are black box with uninterpretable failure modes and hallucinations. Their details, relationships to BDT, and extension to Vision-Language-Action are beyond the scope of this article.

**Generative Models of Images**

Researchers developed network architectures for generating images so realistic that they have won prizes in art competitions. Generative Adversarial Networks (GANs) (Goodfellow et al, 2014) trained a generative network, essentially an upside-down CNN, to generate realistic synthetic images by competing with a network which was trained to discriminate between real images and synthetic ones. The dominant current approaches are Diffusion Models which are trained by an auto-encoder strategy where an image is diffused to output latent variables, and a network is trained to reconstruct the image (Saharia et al. 2022). These generative models can be primed by text captions, using the attention techniques pioneered by transformers and produce an enormous variety of detailed and diverse images. They are very exciting from our Bayesian perspective because they, in combining with the computer graphics techniques used in game engines, suggest the practicality of developing the probability distributions for generating even the most complex images, as required by Bayes.

**Controversies and Conjectures**

Computer vision is a very dynamic research field whose scope is expanding to include interactions with language, reasoning, and much of AI and high-level cognition. It is a major industry with enormous potential. But, as with AI, there are political concerns that it relies too much on black-box techniques which are complex and hard to interpret. It is argued that CV and AI should be responsible with understandable failure modes and performance guarantees and be controllable.

The BDT perspective enables us to contrast between feedforward neural network algorithms which are heavily data driven and the alternative Bayesian perspective. We conjecture that Bayes, through technically more challenging, will be necessary to achieve responsible AI. It enables a duality between analysis and synthesis and the use of structured representations, and 3D world models of objects, agents, their interactions, and knowledge about the world. This is a good fit to theories of human cognition where children behave like "baby scientists" who learn to model the world by making predictions and testing them by experiments.

We conjecture that CV will increasingly influence cognitive science studies of vision. CV provides a rich set of tools for constructing computational models of vision on realistic visual stimuli and can be tested by behavioral experiments. Ideal Observer studies will give insight into human visual abilities which, in turn, may improve CV. CV and AI techniques are already being used to predict fMRI and EEG activity and give understanding to the functions of different visual areas and the relationship between them. Similarly feedforward networks, originally inspired by neuroscience, are increasingly used to make predictions for multi-electrode recordings and address the neural basis of cognition which may, in turn, result not only in better scientific understanding but also to novel AI neural architectures.

**Additional Reading**

Computer vision is a vast field and this article has restricted itself to a limited range of topics viewed from the perspective of BDT. For an alternative, accessible, and up to date overview of many aspects of computer vision we recommend (Torrabla et al. 2024).